\documentclass[10pt,twocolumn,letterpaper]{article}

\usepackage{cvpr}
\usepackage{times}
\usepackage{epsfig}
\usepackage{graphicx}
\usepackage{amsmath}
\usepackage{amssymb}
\usepackage{graphics}
\usepackage{booktabs}
\usepackage{multirow}
\usepackage{enumitem}
\usepackage{bm}
\usepackage[dvipsnames,table]{xcolor}
\usepackage{adjustbox}
\usepackage[innercaption]{sidecap}
\usepackage{upquote}
\usepackage{textcomp}
\usepackage{pifont}
\usepackage{natbib}
\usepackage{etoolbox}
\usepackage{caption}
\usepackage{subcaption}
\usepackage[utf8]{inputenc}
\usepackage{xspace}
\usepackage{enumitem}
\usepackage{wasysym}
\usepackage{lipsum}
\usepackage{pict2e}
\usepackage{relsize}
\usepackage{soul}

\setcitestyle{numbers,square,citesep={,}}
\sidecaptionvpos{figure}{m}
\graphicspath{{./}}
\DeclareGraphicsExtensions{.jpeg,.png,.eps,.pdf}
\newcommand{\Ree}{\mathbb{R}}

\newcommand{\cmark}{\ding{51}}%

\newcommand{\partitle}[1]{\noindent\textbf{#1}}
\newcommand{\ptspace}{\vspace*{5pt}}

\newenvironment{boxedcomment}
{\begin{center}
		\begin{tabular}{|p{0.9\linewidth}|}
			\hline\\
		}
		{ 
			\\\\\hline
		\end{tabular} 
	\end{center}
}

\DeclareFontFamily{U}{mathb}{\hyphenchar\font45}
\DeclareFontShape{U}{mathb}{m}{n}{
	<-6> mathb5
	<6-7> mathb6
	<7-8> mathb7
	<8-9> mathb8
	<9-10> mathb9
	<10-12> mathb10
	<12-> mathb12}{}
\DeclareSymbolFont{mathb}{U}{mathb}{m}{n}
\DeclareMathSymbol{\llcurly}{\mathrel}{mathb}{"CE}
\DeclareMathSymbol{\ggcurly}{\mathrel}{mathb}{"CF}

\definecolor{actioncolor01}{HTML}{953735}
\definecolor{actioncolor02}{HTML}{997800}
\definecolor{actioncolor03}{HTML}{4F6228}
\definecolor{actioncolor04}{HTML}{9B540D}
\definecolor{actioncolor05}{HTML}{376092}
\definecolor{actioncolor06}{HTML}{604A7B}

\definecolor{actioncolor11}{HTML}{F2DCDB}
\definecolor{actioncolor12}{HTML}{FFEDB4}
\definecolor{actioncolor13}{HTML}{D7E4BD}
\definecolor{actioncolor14}{HTML}{FFDFB5}
\definecolor{actioncolor15}{HTML}{DCE6F2}
\definecolor{actioncolor16}{HTML}{E6E0EC}

\DeclareRobustCommand\ActionCircleA{\put(0,2.5){\color{actioncolor01}\circle{7.0}}\put(0,2.5){\color{actioncolor11}\circle*{6.0}}}
\DeclareRobustCommand\ActionCircleB{\put(0,2.5){\color{actioncolor02}\circle{7.0}}\put(0,2.5){\color{actioncolor12}\circle*{6.0}}}

\DeclareRobustCommand\ActionCircleE{\put(0,2.5){\color{actioncolor05}\circle{7.0}}\put(0,2.5){\color{actioncolor15}\circle*{6.0}}}

\definecolor{lightgray}{gray}{0.6}
\cvprfinalcopy

\ifcvprfinal\fi

\begin{document}

\title{PIC: Permutation Invariant Convolution for Recognizing Long-range Activities}

\author{
Noureldien~Hussein, Efstratios~Gavves, Arnold~W.M.~Smeulders
\\
QUVA~Lab, University~of~Amsterdam
\\
{\tt\small\{nhussein, egavves, a.w.m.smeulders\}@uva.nl}}

\maketitle

\begin{abstract}
Neural operations as convolutions, self-attention, and vector aggregation are the go-to choices for recognizing short-range actions.
However, they have three limitations in modeling long-range activities.
This paper presents PIC, Permutation Invariant Convolution, a novel neural layer to model the temporal structure of long-range activities.
It has three desirable properties.
i. Unlike standard convolution, PIC is invariant to the temporal permutations of features within its receptive field, qualifying it to model the weak temporal structures.
ii. Different from vector aggregation, PIC respects local connectivity, enabling it to learn long-range temporal abstractions using cascaded layers.
iii. In contrast to self-attention, PIC uses shared weights, making it more capable of detecting the most discriminant visual evidence across long and noisy videos.
We study the three properties of PIC and demonstrate its effectiveness in recognizing the long-range activities of Charades, Breakfast, and MultiThumos.
\end{abstract}

\section{Introduction}
\begin{figure}[!ht]
	\begin{center}
		\includegraphics[trim=0mm 6mm 0mm 0mm,width=0.95\linewidth]{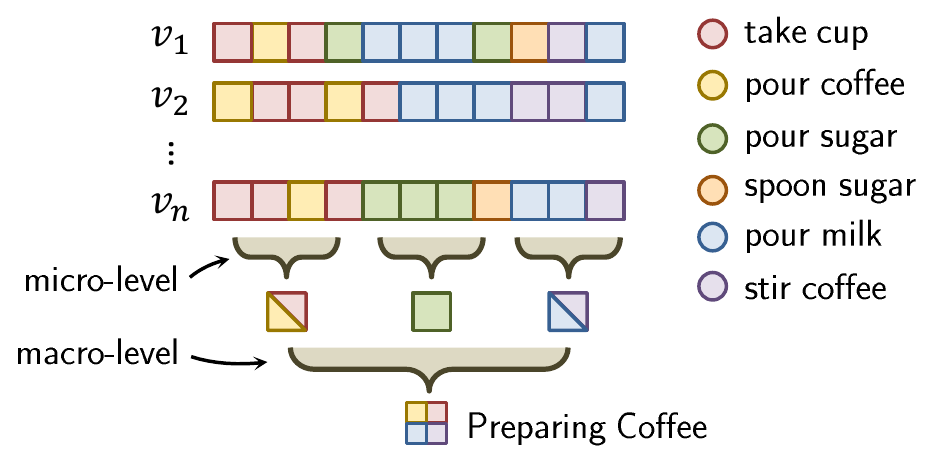}
	\end{center}
	\caption{
		PIC, Permutation Invariant Convolution, recognizes long-range activities using multiple levels of abstractions.
		On the micro-level of a short segment $s_1$, PIC models the correlation between unit-actions, regardless of their order, repetition or duration, 
		$s_1$$=$$\{ \,\,
		\text{\ActionCircleB} \,\,, \,\, \text{\ActionCircleA}  \,\, \} = \{ \,\,
		\text{\ActionCircleA} \,\,, \,\, \text{\ActionCircleB}  \,\,, \,\, \text{\ActionCircleA} \,\, \}$.
		On the macro-level, PIC learns the interactions between segments.
	}
	\label{fig:1-1}
	\vspace*{-5mm}
\end{figure}

Long-range human activities are well-known for being lengthy in duration~\cite{kuehne2014language}, diverse in composition~\cite{hussein2018timeception}, and chaotic in temporal order~\cite{hussein2019videograph}, take for example ``preparing coffee", see figure~\ref{fig:1-1}.
It can take up to ten minutes to unfold, is composed of short, yet plenty, building blocks, called unit-actions~\cite{kuehne2014language}, as ``take cup" and ``pour milk".
Moreover, their temporal order can be very chaotic and unpredictable.
By how many ways one can make a cup of coffee?
And by which specific order of unit-actions?
Is there an overarching structure?

Long-range activities exhibit a temporal structure, albeit complex~\cite{hussein2018timeception}, and have a temporal order, yet weak~\cite{hussein2019videograph}.
Take for example the activity of ``preparing coffee".
Loosely speaking, its temporal structure is analogous to partially ordered sets~\cite{dushnik1941partially}, figure~\ref{fig:1-1}.
This structure can be captured by multiple levels of abstractions.
On the macro-level, the activity is divided into a few segments $v$$=$$\{s_1, s_2, ..., s_n\}$.
On the micro-level, each segment consists of a few but highly correlated unit-action that usually occur within the same neighborhood.
For example, one segment contain the unit-actions $s_1$$=$$\{ \text{``spoon sugar"}, \text{``pour milk"} \}$, while another comprise $s_2$$=$$\{ \text{``take cup"}, \text{``pour coffee"} \}$.
Across video exemplars of preparing coffee, there is no single correct order of these unit-actions in each segment.
The question is how to model such a disordered temporal structure?

\begin{figure}[!t]
\begin{minipage}{1.0\linewidth}
\begin{center}
\includegraphics[trim=2mm 4mm 2mm 0mm,width=1.0\linewidth]{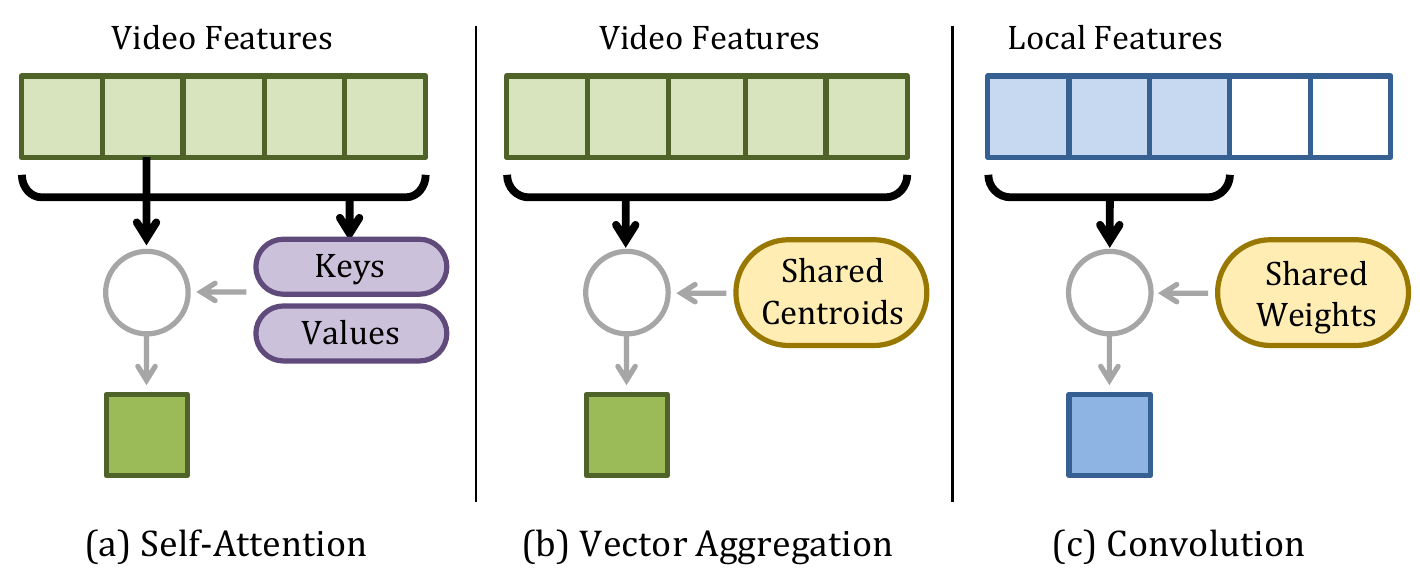}
\end{center}
\vspace*{2pt}
\end{minipage}
\begin{minipage}{1.0\linewidth}
\renewcommand{\arraystretch}{1.2}
\setlength\tabcolsep{9.3pt}
\begin{center}
\begin{tabular}{l|cccc}
& (a) & (b) & (c) & PIC \\
\hline
Permutation Invariance    & \cmark & \cmark &           & \cmark \\
Shared Weights            &        & \cmark & \cmark    & \cmark \\
Temporal locality         &        &        & \cmark    & \cmark \\
\end{tabular}
\end{center}
\vspace*{-5mm}
\end{minipage}
\caption{Compared to other temporal modeling layers, PIC has three benefits.
\textit{i.} Temporal locality to learn long-range temporal abstractions using a cascade of layers.
\textit{ii.} Shared weights (i.e. key-value kernels) to detect the discriminant concepts.
\textit{iii.} Invariant to the temporal permutation with in the receptive field, better for modeling weak structures.}
\label{fig:2-1}
\vspace*{-5mm}
\end{figure}

From the literature, we conclude three predominant approaches for temporal modeling of long-range activities: convolution~\cite{hussein2018timeception,carreira2017quo}, self-attention~\cite{wang2017non,wu2019long} and vector aggregation~\cite{duta2017spatio, girdhar2017actionvlad}, see figure~\ref{fig:2-1}.
Convolution is successful in learning strong temporal patterns~\cite{carreira2017quo} thanks to operating on local windows, {i.e.} convolution regards temporal locality.
Also, it learns long-range dependencies using cascaded layers~\cite{lea2017temporal}.
But even with multi-scale kernels~\cite{hussein2018timeception}, convolution is sensitive to the temporal order of the receptive field, thus it is less suited for modeling the chaotic structure of long-range activities.

In contrast, vector aggregation, is invariant to the temporal order.
But the downside is that it ignores temporal locality.
So, it is unable to learn multiple levels of abstractions using cascaded layers.
Note that there is only one layer used in ActionVlad~\cite{girdhar2017actionvlad}.
Self-attention, as Nonlocal~\cite{wang2017non,wu2019long}, uses key-value pair of vectors to capture the long-range dependencies.
This pair are not shared, but rather are inferred from the input signal~\cite{vaswani2017attention}.
So, self-attention is less successful in detecting the most discriminant visual evidence from the noisy input signal of long-range activities.

To overcome the limitations of previous methods, we propose PIC, Permutation Invariant Convolution, a temporal modeling layer with three novelties:
\textit{i.} Unlike typical convolutions~\cite{carreira2017quo}, PIC is invariant to the temporal permutations within the local window.
The result is being better suited to handle the different temporal orders by which a long-range activity takes place.
\textit{ii.} In contrast to self-attention~\cite{wang2017non}, it uses shared weights for better detection of salient visual evidence in noisy long-range activities.
\textit{iii.} Different from vector aggregation~\cite{duta2017spatio,girdhar2017actionvlad} and self-attention~\cite{wang2017non}, PIC considers local connectivity.
Thus, it learns multiple levels of abstractions using layer cascade.
The outcome of the proposed PIC layer is enabling off-the-shelf CNNs to recognize long-range human activities and outperform existing methods on three Benchmarks: Charades~\cite{sigurdsson2016hollywood}, Breakfast~\cite{kuehne2014language}, and MultiThumos~\cite{yeung2015every}.

\section{Related Work}\label{sec:related_work}

\partitle{Short-range Activities.}
An important task in understanding is recognizing short- and mid-range actions.
These actions usually take up to 10 seconds to occur.
For example, actions is sports, as UCF~\cite{soomro2012ucf101}, Sports-1M~\cite{karpathy2014large}, or human interactions as Kinetics~\cite{kay2017kinetics}.
To address these benchmarks, literature propose methods for modeling the pattern, structure~\cite{lea2017temporal}, order~\cite{ghodrati2018video}, and motion~\cite{wang2011action,jain2013better} of the temporal signals.

\ptspace\partitle{Long-range Activities.}
Recently, there is a major interest in understanding long-range activities, which brings news challenges.
The reason is that these activities are complex~\cite{hussein2018timeception}, take longer to unfold~\cite{kuehne2014language} and are harder to model their temporal structure~\cite{hussein2019videograph,hussein2017unified}.
New benchmarks are proposed, as Charades~\cite{sigurdsson2016hollywood}, Epic-Kitchens~\cite{damen2018scaling}, Breakfast~\cite{kuehne2014language}, MultiThumos~\cite{yeung2015every,idrees2017thumos}, YouCook~\cite{Zhou2017YouCookIID} or Tasty~\cite{sener2019zero}.

This paper focuses on modeling and recognizing long-range human activities.
After a closer look into the related literature of only long-range modeling, one can conclude the prevalence of three approaches: convolution~\cite{hussein2018timeception}, self-attention~\cite{wang2017non,wu2019long}, and vector aggregation~\cite{girdhar2017actionvlad,duta2017spatio}, see figure~\ref{fig:2-1}.

\ptspace\partitle{Convolution.}
In this vein, convolutional kernels learn to detect patterns within a local window, \textit{i.e.} receptive field.
Then, using a cascade of layers, convolution can learn multiple levels of abstractions~\cite{krizhevsky2014imagenet,simonyan2014very}.
So, one can simply attribute the success of convolutional models to two factors: respecting temporal locality and learning complex representations using cascaded layers.
The outcome is many successful CNN architectures for image~\cite{krizhevsky2014imagenet,he2016deep} and action understanding~\cite{ji20123d,simonyan2014two}, temporal localization~\cite{duta2017spatio}, and sequential modeling~\cite{gehring2017convolutional}.

However, temporal convolutions are sensitive to the temporal order, even with multi-scale kernels~\cite{hussein2018timeception,szegedy2017inception}.
Differently, this paper proposes PIC, which is invariant to temporal permutation and more permissive to the many temporal configurations exhibited by a long-range activity.

\ptspace\partitle{Self-attention.}
Attention is extensively used in many tasks as image captioning~\cite{xu2015show}, temporal detection~\cite{sharma2015action} and action recognition~\cite{du2018interaction,li2018videolstm}.
Recently, self-attention shows success in machine translation~\cite{vaswani2017attention} thanks to using a pair of vectors, namely key-value.
Self-attention is adopted by various methods for graph representation~\cite{velivckovic2017graph}, image recognition~\cite{wang2017non}, video understanding~\cite{wu2019long,girdhar2019video}, and efficient recognition~\cite{hussein2020timegate}.

Though, the limitation of self-attention~\cite{wang2017non} is twofold.
First, it ignores temporal locality, which is fundamental to learning multiple levels of abstractions~\cite{parmar2019stand}.
Second, the key-value pairs are inferred from the input, which limits their recognition ability~\cite{lample2019large}.
In contrast, PIC uses weight sharing of the key-value pairs for a better filtering of the visual evidence in a noisy and long activity.
Weight sharing is paramount to not only convolution but also to self-attention, as explained by~\cite{lample2019large}.

\ptspace\partitle{Vector Aggregation.}
This line of work pool feature representations of video frames over long-range sequence. Simple pooling methods is used as max~\cite{hussein2017unified}, attention~\cite{girdhar2017attentional}, and gating~\cite{miech2017learnable}.
While others opt for more complex aggregation as Vlad~\cite{duta2017spatio,girdhar2017actionvlad} and Fisher Vectors~\cite{oneata2013action}.
The upside of such methods is scaling up to long-range activities and being invariant to their scale, order and repetition.

Nevertheless, the downside of vector aggregation is that the temporal locality is ignored, and the temporal structure is overlooked.
That's why Vlad methods opt for only one layer of temporal modeling.
As an alternative, PIC regards temporal locality, thus able to learn multiple levels of abstractions using cascaded layers.

\section{Method}\label{sec:method}

First, we introduce PIC, Permutation Invariant Convolution, and discuss its novelties over existing layers for temporal modeling: convolution~\cite{hussein2018timeception}, self-attention~\cite{wang2017non} and vector aggregation~\cite{girdhar2017actionvlad}.
Then, we describe how it can be fitted on top of modern CNNs.
Finally, we detail the final model architecture and its implementation.

\subsection{Motivation}
The structure of the long-range activities can be thought of as partially ordered sets, which constitute multiple levels of abstractions.
On the macro-level, the entire video $v$ of long-rage activity consists of a few segments $v$$=$$\{ s_1, s_2, s_3, ... \}$,
But on the micro-level, each segment $s_i$ consists a few highly-correlated unit-actions, albeit with no particular order or number of repetitions.
Take for example the activity of ``preparing coffee", see figure~\ref{fig:1-1}.
One segment contains the unit-actions $s_1$$=$$\{  \text{``take cup"},  \text{``pour coffe"} \}$, while another comprise $s_2$$=$$\{ \text{``pour sugar"}, \text{``spoon sugar"},  \text{``pour milk"} \}$, and so on so forth.
It is demonstrated by~\cite{hussein2018timeception} that the multi-level structure of long-range activity can be learned using convolutional approach with a cascade of layers.
The bottom layers learn the correlation between the unit-actions within each segment, while the top layers learn the interactions between the segments.
First, we discuss the standard temporal convolution, and its limitation in modeling the chaotic structure of long-range activities.

\partitle{Standard Temporal Convolution.}
As we are interested in temporal modeling, we omit the spatial dimensions and focus only on the temporal dimension, for clarity.
For which, the temporal convolution works as follows.
It relies on a learned kernel $W = \{w_i \; | \; i \in [1, ..., T] \}, W \in \Ree^{T \times C}$, where $T, C$ are the kernel size and  dimension, respectively.
At the $i$-th timestep, the input feature in a local window $X_w = \{ x_i \; | \; i \in [1, ..., T] \}$ is convoluted $(\circledast)$ with the kernel $W$, the output feature is $y \in \Ree^{1 \times 1}$.
So, standard temporal convolution is formulated as
\begin{equation}
\begin{split}
y &= W \circledast X_w = \sum_{i=1}^{T} w_i \odot x_i^{\top}.
\end{split}
\end{equation}
With this operation, the kernel $W$ learns to detect the exact temporal order of the sequence $X_w$.
However, the downside is that this operation is sensitive to the precise sequential order of $X_w$.
In other words, standard temporal convolutions are not permissive to the many temporal configurations a sequence of unit-actions can take place in a long-range activity.
For example, there is no one particular order by which the sequence $ \{ \text{``pour sugar"}, \text{``take cup"}, \text{``pour milk"} \}$ can occur in the activity of ``preparing coffee".
One possible solution is multi-scale convolutions~\cite{hussein2018timeception}.
They can model temporal sequences that differ in their temporal extent.
However, they are still sensitive to the temporal order.
Another possible solution is using more convolutional kernels, such that each learns a different temporal order.
This solution is computationally prohibitive, and cannot account for all possible permutations, especially for longer temporal patterns.

So, to successfully model long-range activities, a strong requirement is that the convolution operation has to be invariant to the temporal order of unit-actions within the local window, \textit{i.e.} within the receptive field.
For there exist many ways one can perform the activity of ``preparing coffee", with no strict order.
To this end, we propose PIC, an invariant convolutional operation to replace the standard convolution for temporal modeling of long-range activities.

\subsection{PIC: Permutation Invariant Convolution}

\begin{figure}[!t]
	\begin{center}
		\includegraphics[trim=0mm 8mm 0mm 4mm,width=0.9\linewidth]{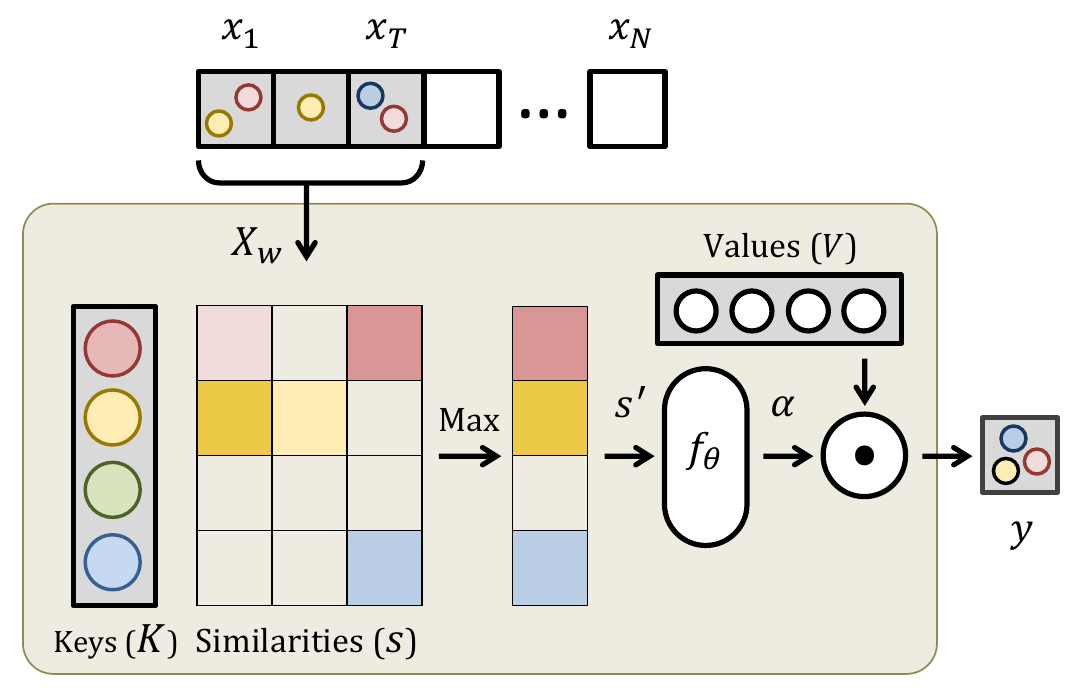}
	\end{center}
	\caption{Overview of PIC, Permutation Invariant Convolution. Using a pair of Key-Value kernels $(K, V)$, it models the correlation between the visual evidences $\{  \; \textbf{\ActionCircleA} \;, \; \textbf{\ActionCircleB} \;, \; \textbf{\ActionCircleE} \; \}$ in a local window with $X_w = \{ x_1, ..., x_T \}$ irrespective of their the temporal order.}
	\label{fig:3-1}
	\vspace*{-5mm}
\end{figure}

The goal is to make the standard convolution permissive to the weak temporal order of long-range activities.
We propose PIC, Permutation Invariant Convolution, see figure~\ref{fig:3-1}.
PIC takes as an input the features $X_w$ in a local window.
To model their correlations regardless of their order, PIC uses a pair of kernels, inspired by self-attention operation~\cite{wang2017non,vaswani2017attention}.
The pair is demoted as the keys $K \in \Ree^{M \times C}$ and the values $V \in \Ree^{M \times C}$, where $M$ is the number of kernels, and $C$ is the kernel dimension.
The keys $K$ are known to act as a detector for $M$ \textit{latent} visual concepts.
Using an outer product $\otimes$ between the keys $K$ and all the features of the local window $X_w$, we get the similarity matrix $s \in \Ree^{M \times T}$.
Intuitively, $s$ encodes the possibility of any of the $M$ latent concepts to ever exist in the local window.
By max-pooling the similarities $s$ over the temporal dimension of the local window, we get $s^{\prime} \in \Ree^{M \times 1}$.
We interpret $s^{\prime}$ as the maximum possibility of $M$ concepts to take place in the local window.

After obtaining the maximum similarities $s^{\prime}$, we opt for values kernel $V$ to represent only those detected.
The main purpose of using a pair of kernels $K ,V$ instead of one is twofold.
First, using a pair enables PIC to decouple detecting the concepts using the keys $K$, from representing them with the values $V$.
Decoupling is proposed by~\cite{vaswani2017attention} and successfully used in~\cite{wang2017non}.
Second, by decoupling the kernels, we can have more keys $K \in \Ree^{M\times C}$ for detection and less values $V \in \Ree^{M^{\prime} \times C}$ for representation, where $M^{\prime} << M$.

\begin{figure}[!ht]
	\begin{center}
		\includegraphics[trim=0mm 8mm 0mm 5mm,width=0.5\linewidth]{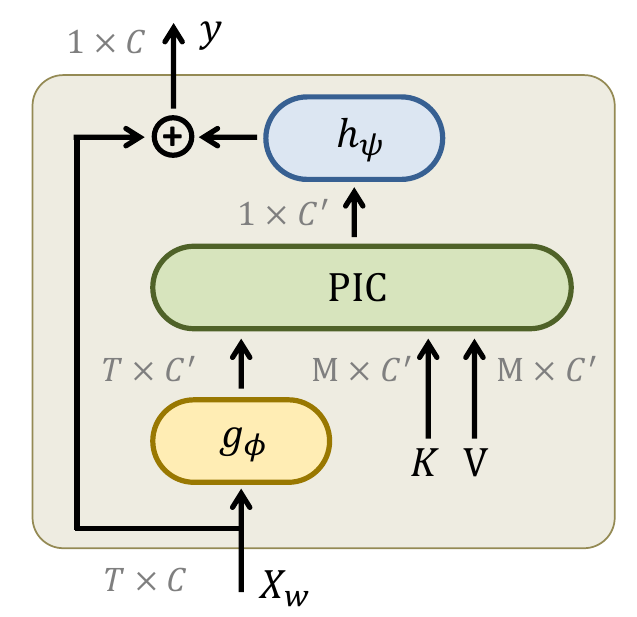}
	\end{center}
	\caption{PIC layer models only the temporal dimension.
		It has shared kernels $K,V$ to learn discriminant concepts.
		A residual bottleneck is used to reduce computation.}
	\label{fig:3-2}
	\vspace*{-5mm}
\end{figure}

The next step is using a dense layer $f_{\theta}(\cdot)$ to model the correlation between the maximum similarities $s^{\prime}$, and also to embed them from a higher dimension $\Ree^{M \times 1}$ to a lower dimension $\Ree^{M^{\prime} \times 1}$.
Then, an activation $\sigma = \texttt{ReLU}$ is used to rectify the similarities, resulting in the activated similarity $\alpha \in \Ree ^{M \times 1}$.
The final step is an inner product $\odot$ between the similarities $\alpha$ and the values $V$ to arrive at the final representation $y \in \Ree ^{1 \times C}$.
PIC is formulated as
\begingroup
\allowdisplaybreaks
\begin{align}
s &= K \otimes X_w^{\top} \label{eqn:3-2-1}
\\
s^{\prime} &= \max \vspace{0pt} _{\mathrm{row}} (s) \label{eqn:3-2-2}
\\
\alpha &= \sigma \left[ f_{\theta} (s^{\prime}) \right]
\\
y &= \alpha^{\top} \odot V.
\end{align}
\endgroup
\partitle{PIC Layer.}
After outlining the PIC operation, now we discuss how PIC can be used as a modular layer.
PIC is a convolutional neural layer placed on top of backbone CNNs -- be it 2D or 3D, see figure~\ref{fig:3-2}.
It draws inspirations and design principles from a few related works~\cite{hussein2018timeception,wang2017non,xie2017aggregated}.
In total, we list four design principles that govern PIC layer.
\textit{i.} PIC uses a residual bottleneck for reducing the computation~\cite{he2016deep,xie2017aggregated}.
Before convolving the input features $X_w \in \Ree^{T \times C}$ with PIC,
their dimension is reduced from $C$ to $C^{\prime} = C / 4$ using a dense layer $g_\phi (\cdot)$.
And to enable residual connection, the input dimension $C$ is recovered by another dense layer $h_\psi (\cdot)$.
\textit{ii.} Instead of using one kernel as in standard convolution, PIC uses a pair of key-value kernels $(K, V)$~\cite{wang2017non,vaswani2017attention}, to decouple concept detection from concept representation.
\textit{iii.} PIC focuses on modeling only the temporal dimension~\cite{hussein2018timeception}, leaving the spatial dimensions for the backbone CNN to handle.
\textit{iv.} Similar to the kernels of standard convolution, the kernels $K,V$ learned by PIC are shared weights, \textit{i.e.} model parameters, and are not inferred from the window features $X_w$.
While in~\cite{wang2017non}, the keys and values $K, V$ are inferred from the input $X_w$.
The upside of having shared kernels $K, V$ is the ability to detect the most representative visual concepts across the entire long-range activity, and not being conditioned on the visual signals in a narrow local window.
This is an important design choice for modeling such activities, particularly when we do not know if $X_w$ ever contains informative or noisy evidence.
In addition, PIC respects temporal locality. In other words, it convolve the features of local windows $X_w$, in contrast to global windows used in self-attention~\cite{wang2017non}.
temporal locality enables PIC to learn multiple levels of abstractions with cascaded layers.

\subsection{Final Model}
We start with an off-the-shelf backbone CNN, be it 2D CNN as ResNet~\cite{he2016deep} or 3D CNN as I3D~\cite{carreira2017quo}.
Then, we stack a cascade of four PIC layer layers on top of the last convolution layer of the backbone CNN.
Each layer consists of PIC convolution followed by \texttt{BatchNorm} for normalization, \texttt{LeakyReLU} for activation, and \texttt{MaxPool} with stride 2 for downsampling.

Given a video $v$ of long-range activity, we uniformly sample $N$ segments $v = \{s_j \; | \; j \in [1, ...N] \}$.
Each segment $s_j$ consists of $L=8$ successive video frames, and is processed by the backbone CNN, up to the last convolution layer.
The output convolutional feature is $x_j \in \Ree^{1024 \times 7 \times 7}$.
We call $x_j$ the feature of the $j$-th timestep, because it corresponds to the $j$-th segment of the video.
The video-level features are then ${\bm X} = \{ x_j \; | \; j \in \{1, ..., N\}$, where $N$ is the temporal dimension, or the number of timesteps.
To model the temporal structure of the entire video $v$, we feed-forward the features ${\bm X}$ to the cascade of PIC layers.
Thanks to using a downsampling with stride 2, and four PIC layers in the cascade, the temporal footprint of the input features ${\bm X}$ is reduced to $N/4$.
And so, the output feature is $Z \in \Ree^{1024 \times 7 \times 7 \times N/4}$.
For video classification, $Z$ is pooled over the spatial and temporal dimensions, and feed-forwarded to a two-layer MLP with \texttt{BatchNorm} and \texttt{ReLU}.
The MLP uses \texttt{softmax} and \texttt{sigmoid} as the last activation functions for the tasks of single-label and multi-label classification, respectively.

\ptspace\partitle{Implementation.}
For each dataset, we follow a two-stage procedure to train our final model.
In the first stage, the backbone CNN is pre-trained on the dataset at hand.
We follow the same training details provided by the authors of the backbone CNN, for example I3D~\cite{carreira2017quo}.
In the second stage, the cascade of PIC layers is placed on top of the last convolutional layer of the backbone CNN.
Only PIC layers, along with the classifier, are trained on the dataset at hand, while the backbone is kept frozen.
The model is trained for 100 epochs and with batch size 32.
For optimization, we opt for SGD with 0.1, 0.9 and 1e-5 as the learning rate, momentum and weight decay, respectively.
Also, we experiment Adam with 0.01 and 1e-4 as the learning rate and epsilon $\epsilon$, respectively.
TensorFlow~\cite{tensorflow2015-whitepaper} and Keras~\cite{chollet2015keras} are used for implementation. Code is made public upon publication.

\section{Experiments}\label{sec:experiments}
As manifested by figure~\ref{fig:2-1}, there exist three predominant approaches for temporal modeling and recognizing long-range activities.
These approaches: self-attention, vector aggregation and convolution, and they are exemplified by the following temporal layers: Nonlocal~\cite{wang2017non}, and ActionVlad~\cite{girdhar2017actionvlad} and Timeception~\cite{hussein2018timeception}, respectively.
In this section, we compare PIC against these layers.
In addition, we conduct a comprehensive analysis of the properties of PIC and showcase how it enables existing CNNs to better model and recognize the long-range activities.

\subsection{Datasets}
\partitle{Charades}~\cite{sigurdsson2016hollywood}
is video dataset for multi-label action classification, with total number of 157 unit-action classes.
It contains 8k, 1.2k and 2k videos for training, validation and test splits, respectively (67 hrs for training split).
Each video can be thought of a long-range human activity.
On average, each video is 30 seconds and contains 6 different unit-actions.
Thus, Charades meets the criteria of complex actions.
We use mean Average Precision (mAP) for evaluation.

\ptspace\partitle{Breakfast}~\cite{kuehne2014language}
is a dataset for unscripted cooking-oriented human activities.
It contains 1712 videos in total, 1357 for training and 335 for test.
The average length of videos is 2.3 minutes.
It is a video classification task of 10 categories of breakfast activities, where each video represents only one activity.
Besides, each video has 5 unit-actions composing its activity.
In total, there are 48 classes of unit-actions.
In our experiments, we only use the activity annotation, and we do not use the annotation of unit-actions.

\ptspace\partitle{MultiTumos}~\cite{yeung2015every}
is a dataset for untrimmed human activities in videos, with the primary focus of temporal localization.
It contains 65 action classes and 400 videos (30 hrs).
Each video can be thought of a complex action, which comprises 11 unit-actions on average.
MultiThumos extends the original Thumos-14~\cite{idrees2017thumos} by providing multi-label annotation for the videos in validation and test splits.
Having multiple and dense labels for the video frames enable temporal models to benefit from the temporal relations between unit-actions across the video.
The metric mAP is used for evaluation.

\subsection{Dissection of PIC}
As presented earlier, PIC is a convolutional operation better suited for recognizing long-range activities, thanks to three favorable properties: \textit{i.} invariance to permutation, \textit{ii.} respect local connectivity, \textit{iii.} using shared kernels for the key-value pairs.
So, in the following experiments, we dissect the PIC layer to highlight the individual importance of each of these three properties.
These experiments use Breakfast~\cite{yeung2015every}, as it is the only available dataset for single-label recognition of long-range activities.

\ptspace\partitle{Permutation Invariance.}
PIC is, by design, invariant to the temporal permutations within windows of local connectivity.
It achieves so by two operations: outer product $\otimes$ between the input $X_w$ and the keys $K$ in equation~\ref{eqn:3-2-1}, and the $\max \vspace{0pt} _{\mathrm{row}}(\cdot)$ operation in equation~\ref{eqn:3-2-2}.
To examine the importance of invariance, we build a variant of PIC, named PIC-Ordered.
In which, we convolve $\circledast$ the input $X_w$ with $K$ instead of using outer product $\otimes$.
And we remove the $\max \vspace{0pt} _{\mathrm{row}}(\cdot)$ operation, thus making PIC-Ordered dependable on the temporal order within the local window $X_w$.
PIC-Ordered is formalized as
\begin{equation}
\begin{split}
\alpha = K \circledast X_w^{\top},\;\;\;\;\; y = \alpha^{\top} \odot V.
\end{split}
\end{equation}
Then, we train baselines accordingly and measure the performance.
Timeception is included in this comparison, as it is a multi-scale convolutional layer, and able to handle slight temporal permutations.

\begin{figure}[!ht]
	\begin{center}
		\includegraphics[trim=0mm 5mm 0mm 2mm,width=0.85\linewidth]{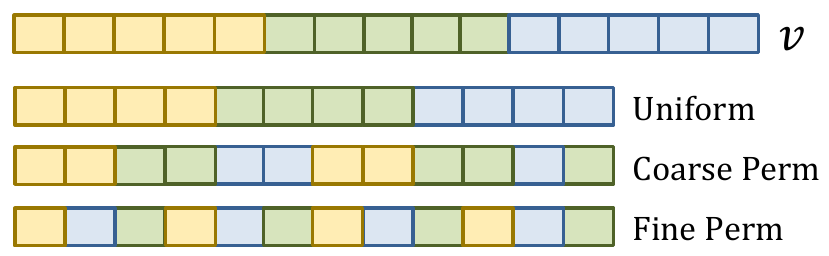}
	\end{center}
	\caption{
		Three different ways of sampling timesteps from a test video: uniform, coarse, and fine permutation.}
	\label{fig:4-1}
	\vspace*{-2mm}
\end{figure}

During testing, we use three different ways to sample $N$ timesteps from a test video: \textit{i} uniform, \textit{ii.} coarse permutation, and \textit{iii.} fine permutation, see figure~\ref{fig:4-1}.
The reason is that we want to introduce randomness to the temporal order, and measure how the baseline methods perform in such cases.
Results are reported in table~\ref{tbl:4-4}.

\begin{table}[!ht]
	\centering
	\renewcommand{\arraystretch}{1.0}
	\setlength\tabcolsep{5.5pt}
	\begin{tabular}{lcccc}
		\specialrule{0.4mm}{.0em}{.3em}
		Baseline  & Coarse Perm. & $\downarrow$ & Fine Perm. & $\downarrow$ \\
		\midrule
		Timeception    &  84.6  $\rightarrow$ 82.2  & 2.4  &  84.6 $\rightarrow$ 81.9  & 2.7 \\
		PIC-Ordered    &  80.2  $\rightarrow$ 77.6 & 2.6   &   80.2 $\rightarrow$ 76.3  & 3.9  \\
		\midrule
		\textbf{PIC}       &  87.5  $\rightarrow$ 87.0 & \textbf{0.5} &  87.5 $\rightarrow$ 86.7  & \textbf{0.8}  \\
		\specialrule{0.4mm}{.0em}{.0em}
	\end{tabular}
	\caption{Being invariant to permutations, PIC is affected the least by altering the temporal order of test videos.}
	\label{tbl:4-4}
	\vspace*{-5pt}
\end{table}

Our observation is that not only PIC outperforms other layers, but also is has the lowest drop in performance in both cases of fine and coarse perturbations of the temporal information.
In addition, we notice that Timeception is slightly more tolerant to perturbations, thanks to its multi-scale kernels.
The conclusion is that PIC is more permissible than others to the many ways a long-range activity can happen.

\ptspace\partitle{Local \textit{v.s.} Global Connectivity.}
PIC is a convolution layer with a temporal receptive field of size $T$.
That's to say, given $N$ features corresponding to $N$ timesteps of a video, PIC operates on local windows, each of size $T$, where $T \ll N $.
This gives PIC the ability to learn temporal abstractions of long-range activities at different layers of the network.
Our assumption is, if a temporal layer is globally connected to the entire video, then there is no need to cascade multiple layers, as this layer would already summarize all the visual evidence in this video.
Note that local connectivity is fundamental to convolutions as well as self-attention~\cite{parmar2019stand}.
To test this assumption, we devise a variant of PIC, called PIC-Global, that is not restricted by a window size.
Its receptive field is as big as the input video $T$$=$$N$.
Then, we train baselines fitted with PIC-Global and PIC.
In this comparison, we include ActionVlad and Nonlocal, as both are temporal layers with global receptive field.

\begin{table}[!ht]
	\centering
	\renewcommand{\arraystretch}{1.0}
	\setlength\tabcolsep{9.0pt}
	\begin{tabular}{lcccc}
		\specialrule{0.4mm}{.0em}{.3em}
		\multirow{2}{*}{Baseline} & \multicolumn{4}{c}{Accuracy (\%) @ Layer} \\ 
		\cmidrule(lr){2-5}
		& 1 & 2  & 3 & 4  \\
		\midrule
		ActionVlad    &  83.07  & --- &  --- &  --- \\
		Nonlocal      &  82.29  & \underline{83.33} &  83.07  & 83.29 \\
		PIC-Global    & \underline{86.76} & 85.68 &  85.68  & 85.42 \\
		\midrule
		Timeception   &83.85,   &    84.90 &   85.30 & 86.93 \\
		PIC      & 86.20  & 87.72 &  88.02  & \textbf{89.84} \\
		\specialrule{0.4mm}{.0em}{.0em}
	\end{tabular}
	\caption{Having a local receptive field enables PIC to learn levels of abstractions at multiple layers.
		Thus, improving monotonically by stacking more layers.
		Others don't witness the same benefit, as they use global receptive field.}
	\label{tbl:4-5}
	\vspace*{-5pt}
\end{table}

As shown in table~\ref{tbl:4-5}, both Timeception and PIC improve monotonically by stacking more layers.
In contrary, the other layers witness a performance plateau after the first or second layer is the stack.
The conclusion is that, the complexity of long-range activities can be captured by a temporal layer of local receptive field.
And over a cascade of layers, the entire complexity is learned.

\ptspace\partitle{Shared \textit{v.s.} Inferred Kernels.}
Inspired by the self-attention~\cite{wang2017non,vaswani2017attention}, PIC uses a pair of kernels $K, V$ to learn latent concepts.
But the difference is that, in PIC, these kernels are shared weights, and not inferred from the input video as in~\cite{wang2017non}.
In short-range videos, it is acceptable to have $K,V$ inferred from a sampled segment from the video, it usually contains most, if not all of the representative visual evidence.
But in long-range video, the sampled segment might not contain all the evidences.
To verify the importance of shared kernels, we construct a variant, named PIC-Inferred.
In which, the pair $K, V$ are inferred from the input features $X_w$ using two dense layers $g_\gamma(\cdot), g_\lambda(\cdot)$, similar to Nonlocal~\cite{wang2017non}.
It is formulated as
\begin{equation}
\begin{split}
K = g_\gamma \left(X_w \right),\;\;\;\;\; V = g_\lambda \left(X_w \right).
\end{split}
\end{equation}
Then, we train baselines and compare their results.
We include Nonlocal in this comparison, as it also uses inferred kernels $K, V$.
The outcome is reported in table~\ref{tbl:4-6}.
We observe that PIC outperforms the other baselines by a considerable margin.
The conclusion is, when it comes to modeling the long-range activities, its important for the convolutional temporal layers to use shared kernels.

\begin{table}[!ht]
	\centering
	\renewcommand{\arraystretch}{1.0}
	\setlength\tabcolsep{9.0pt}
	\begin{tabular}{lcccc}
		\specialrule{0.4mm}{.0em}{.3em}
		\multirow{2}{*}{Baseline} & \multicolumn{4}{c}{Accuracy (\%) @ Layer} \\ 
		\cmidrule(lr){2-5}
		& 1 & 2  & 3 & 4  \\
		\midrule
		Nonlocal      &  82.29  & \underline{83.33} &  83.07  & 83.29 \\
		PIC-Inferred  & 82.55  & 83.85  & \underline{84.90} & 84.64  \\ 
		\midrule
		PIC      & 86.20  & 87.72 &  88.02  & \textbf{89.84} \\
		\specialrule{0.4mm}{.0em}{.0em}
	\end{tabular}
	\caption{Thanks to sharing the kernels $(K, V)$, PIC is better at learning concepts than layers with inferred kernels.}
	\label{tbl:4-6}
	\vspace*{-5pt}
\end{table}

\begin{figure*}[!ht]
	\begin{center}
		\includegraphics[trim=5mm 13mm 5mm 5mm,width=1.0\linewidth]{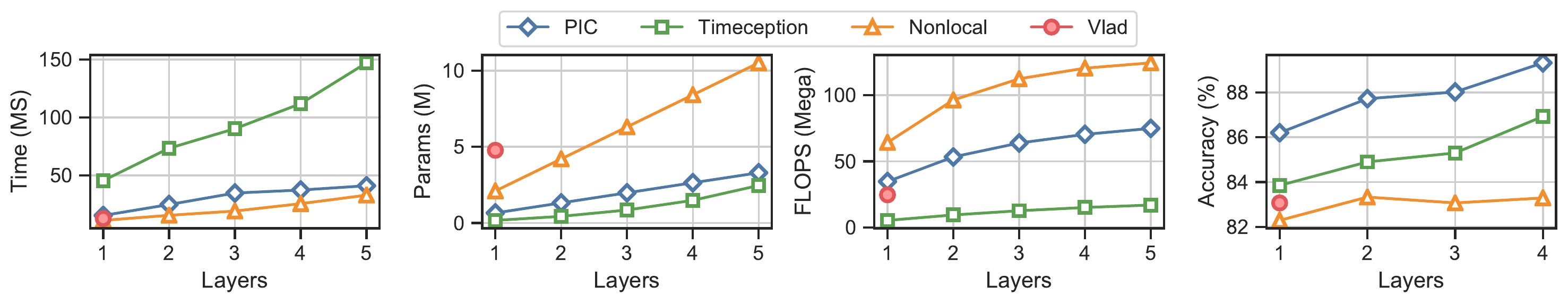}
	\end{center}
	\caption{
		On x-axis, the number of stacked layers.
		While on y-axes, the the efficiency of temporal layers using four metrics: \textit{i.} CPU feedforward time (milliseconds), \textit{ii.} model parameters (millions), \textit{iii.} number of operations (mega FLOPS), and \textit{iv.} classification accuracy (\%).
		PIC has the best tradeoff between efficiency and effectiveness.}
	\label{fig:4-2}
	\vspace*{-5mm}
\end{figure*}

\subsection{Analysis of PIC}
PIC in a modular temporal layer that resides on top of existing backbone CNNs -- be it 2D as ResNet or 3D as I3D.
To better utilize it for these CNNs, we analyze the upsides and downsides of PIC.
And we study three factors: \textit{i.} effectiveness \textit{v.s.} efficiency., \textit{ii.} optimal sizes of receptive field and downsampling, \textit{iii.} extensibility to input video length, and \textit{iv.} scalability with backbone CNNs.

\ptspace\partitle{Effectiveness \textit{v.s.} Efficiency.}
In this analysis, we demonstrate that PIC is an efficient layer for temporal modeling.
Also, we show that PIC scales sub-linearly using deeply cascaded layers.
We compare against other layers for temporal modeling.
Most notably, we include Timeception~\cite{hussein2018timeception}, as it is known for its efficiency.
When quantifying the efficiency, we use four metrics: \textit{i.} CPU feedforward time in milliseconds, \textit{ii.} number of model parameters in millions, \textit{iii.} number of floating point operations in mega FLOPs, and \textit{iv.} classification accuracy of Breakfast activities.

As shown in figure~\ref{fig:4-2}, PIC is very efficient layer, and it scales sub-linearly when stacked.
One observation is that Timeception and PIC are the most efficient layers, and both brings about monotonic improvements in the accuracy using cascaded layers.
Nevertheless, PIC outperforms Timeception by a considerable margin.
We conclude from this analysis that PIC achieves the best tradeoff between efficiency and effectiveness.

\ptspace\partitle{Size of Receptive Field and Downsampling.}
PIC is, in principle, a convolutional operation applied to windows of local connectivity along the temporal dimension of long-range activities.
As such, two of its most important hyperparameters are the window size and downsampling size.
Here, we experiment different configurations to arrive at the best choice.
For this, we use two layers of PIC, each of window size $T$ and followed by a max-pooling operation for downsampling, with stride $s$.

Our observation is that, while increasing the window size helps PIC to have a bigger receptive field, this improvement degrades for $T > 9$.
Based on this analysis of Breakfast dataset, the recommended window size is $T=9$.
As for the downsampling, we find that $s=2$ is the optimal stride, while more aggressive strides $s=\{ 3, 4 \}$ are detrimental.

\begin{table}[!ht]
	\centering
	\renewcommand{\arraystretch}{1.0}
	\setlength\tabcolsep{5.0pt}
	\begin{tabular}{cccccc}
		\specialrule{0.4mm}{.0em}{.3em}
		\multirow{2}{*}{Stride Size ($s$)} & \multicolumn{5}{c}{Window Size ($T$)} \\
		\cmidrule(lr){2-6}
		& 3 & 5 & 7 & 9 & 11 \\
		\midrule
		2     & 86.98  & 86.98 & 87.24 & \textbf{88.02} & 87.50  \\
		3     & 86.98  & 86.72 & \underline{86.98} & 86.20 & 86.20  \\
		4     & 85.68  & 85.68 &  \underline{85.94} & 85.42 & 85.68  \\
		\specialrule{0.4mm}{.0em}{.0em}
	\end{tabular}
	\caption{PIC accuracy when changing the convolution window size $T$ and the downsampling stride size $s$.}
	\label{tbl:4-3}
	\vspace*{-10pt}
\end{table}

\ptspace\partitle{Number of Latent Concepts.}

PIC makes used of shared pair of kernels ($K$, $V$), where $M, M'$ are the number of $K$ and $V$ respectively.
In our experiments, we found that choosing these hyperparameters is of importance to the accuracy.
A rule of thump is, for large datasets, as Charades, where there are many action categories, using large number of keys and values $M = M' = \{ 64, 128\}$ is important.
While in medium-scale datasets, as Breakfast, we found that as little as $M = M' = \{ 16, 32\}$ would suffice.

\subsection{Qualitative Analysis}

\begin{figure*}[!t]
	\begin{center}
		\includegraphics[trim=0mm 8mm 0mm 5mm,width=1.0\linewidth]{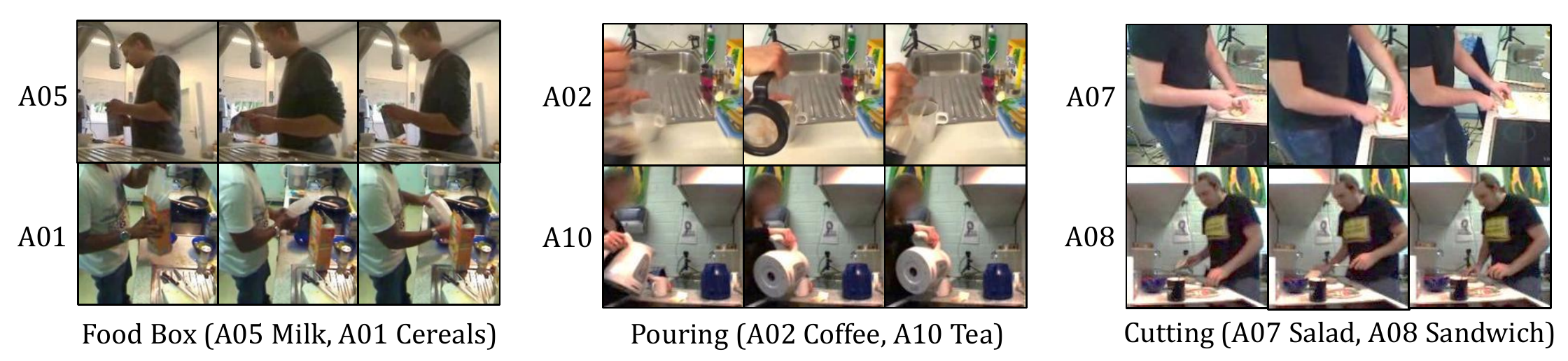}
	\end{center}
	\caption{
		In a cascade of PIC layers, we notice that in the bottom layer, the learned concepts are fine-grained and independent of the activity category.
		For example, the concept ``Pouring" is irrespective of activities ``coffee" or ``tea". Also, these concepts can be object-centric as ``food box" or action-centric as ``cutting".}
	\label{fig:4-4}
	\vspace*{-0mm}
\end{figure*}

\partitle{Learned Concepts.}
PIC learns \textit{latent} concepts using the key kernels $K$.
To interpret what is learned by these kernels, we use Breakfast Activities dataset.
Then, we retrieve the top related video frames to each concept according to the similarity values $s^{\prime}$, see equation ~\ref{eqn:3-2-2}.
What we observe is the following.
In a cascade of PIC layers, we notice that in the bottom layer, the learned concepts are fine-grained and independent of the activity category.
For example, as illustrated in figure~\ref{fig:4-4}, the concept ``pouring" is irrespective of activities ``coffee" or ``tea". Also, these concepts can be object-centic as ``food box" or action-centric as ``cutting".

\begin{figure*}[!t]
	\begin{center}
		\includegraphics[trim=0mm 8mm 0mm 5mm,width=1.0\linewidth]{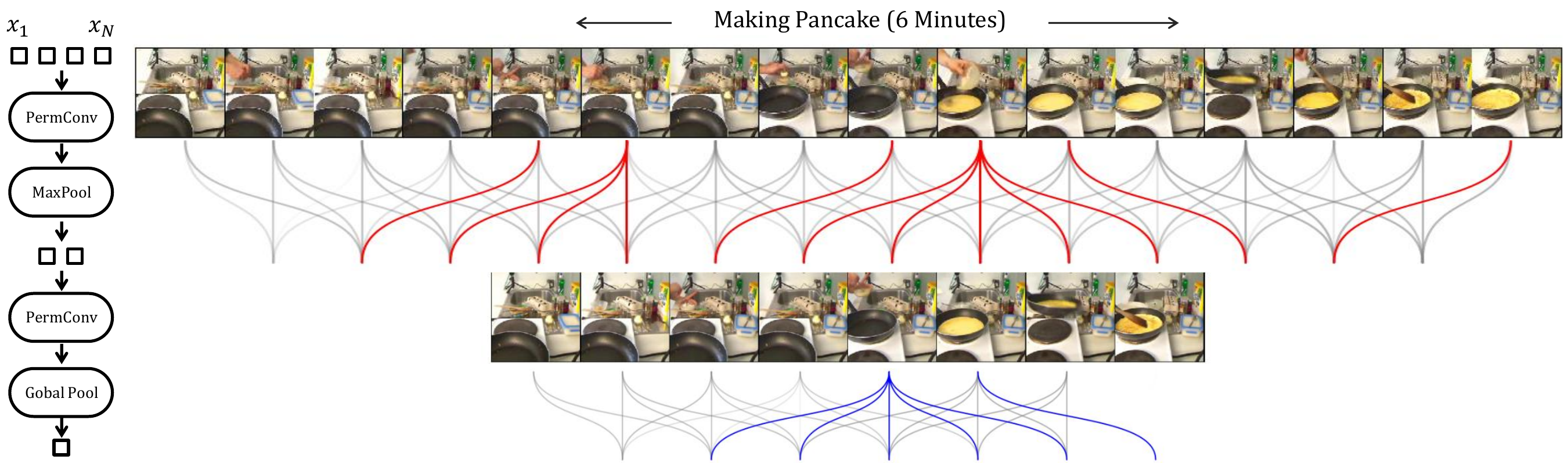}
	\end{center}
	\caption{
		This figure shows 16 frames uniformly sampled from an activity of ``Making Pancake".
		After one layer, $M$ concept kernels are learned to detect relevant visual evidences.
		For simplicity, we show the activations of only one kernel, in red.}
	\label{fig:4-5}
	\vspace*{-5mm}
\end{figure*}

\ptspace\partitle{Long-range Temporal Dependencies.}
The visualization in figure~\ref{fig:4-5} shows a conceptual overview of how a cascade of PIC layers work.
The first layer applies PIC convolution in a sliding window fashion over the temporal dimension.
Each key kernel learns to detect a certain visual evidence.
For simplicity, we show in red the activation of only one kernel.
After the first layer, we down-sample the temporal dimension and follow with another new PIC layer, which by itself learns a new set of kernels.

\subsection{Quantitative Analysis}

\partitle{Breakfast} is the first dataset we use to test our method.
We compare against the three competitive layers for temporal modeling.
In this setup, Timeception uses four stacked layers.
As for Nonlocal, we found that stacking only two layers yielded the best performance.
Lastly, we use four layers of PIC.
All the baselines are trained with the same experimental setup.
The reason our results, reported in table~\ref{tbl:5-1}, are much higher than the results reported in Timeception~\cite{hussein2018timeception} is that we fine-tune the backbone CNN on Breakfast before testing baseline methods.
This paper depends on Breakfast as the main test dataset.
Thus fine-tuning is necessary.

\begin{table}[!ht]
	\centering
	\renewcommand{\arraystretch}{1.0}
	\setlength\tabcolsep{5.0pt}
	\begin{tabular}{lcc}
		\specialrule{0.4mm}{.0em}{.3em}
		Method & Backbone & Accuracy (\%)  \\
		\midrule
		3D CNN & I3D & 80.64 \\ 
		3D CNN + Vlad~\cite{arandjelovic2013all} & I3D & 82.67  \\
		3D CNN + Nonlocal~\cite{wang2017non}  & I3D & 83.79 \\
		3D CNN + Timeception~\cite{hussein2018timeception} & I3D & 86.93 \\
		3D CNN + \textbf{PIC}  & I3D & \textbf{89.84} \\
		\specialrule{0.4mm}{.0em}{.0em}
	\end{tabular}
	\caption{We report the accuracy of classifying the minutes-long activities of Breakfast.
		PIC outperforms the other baseline methods by a considerable margin.}
	\label{tbl:5-1}
	\vspace*{-10pt}
\end{table}

We observe that both Timeception and PIC outperform the other methods, because they learn deep temporal abstractions using cascaded layers.
Nevertheless, PIC outperforms Timeception by a large margin.

\ptspace\partitle{Charades} is used as the third dataset for benchmarking our method.
This dataset is challenging because it is multi-label classification.
A complex action of Charades comprise on average 6 unit-actions.
It is important to mention that since the videos of Charades are noticeably shorter than those of Breakfast, we found that only three layers of PIC cascaded on top of the backbone CNN is optimal.
Stacking the fourth PIC layer does not bring about improvements in this dataset.

\begin{table}[!ht]
	\centering
	\renewcommand{\arraystretch}{1.0}
	\setlength\tabcolsep{13.0pt}
	\begin{tabular}{lcc}
		\specialrule{0.4mm}{.0em}{.3em}
		Method & Backbone & mAP (\%) \\
		\midrule
		SlowFast~\cite{feichtenhofer2019slowfast} & --- & 42.1 \\
		SlowFast-NL$^\ast$~\cite{feichtenhofer2019slowfast} & --- & 42.5 \\
		\midrule
		3D CNN~\cite{wang2017non} & R101  & 35.5 \\
		3D CNN + TC~\cite{hussein2018timeception} & R101 & 41.1 \\
		3D CNN + \textbf{PIC} & R101 & \textbf{42.7} \\
		\midrule
		3D CNN~\cite{wu2019long} & R101-NL & 41.0 \\
		3D CNN + FB~\cite{wu2019long} & R101-NL & 42.5 \\
		3D CNN + \textbf{PIC} & R101-NL & \textbf{43.8} \\
		\specialrule{0.4mm}{.0em}{.0em}
	\end{tabular}
	\caption{When classifying the complex multi-label actions of Charades, PIC layers bring improvements over previous works.}
	\label{tbl:5-2}
	\vspace*{-15pt}
\end{table}

\ptspace\partitle{MultiThumos} is chosen as the third and last dataset to experiment our model.
We follow the same experimental setup as suggested by~\cite{hussein2018timeception} and we use their backbone CNN without fine-tuning on MultiThumos.
This makes the results reported in table~\ref{tbl:4-7} comparable with the results of~\cite{hussein2018timeception}.

\begin{table}[!ht]
	\centering
	\renewcommand{\arraystretch}{1.0}
	\setlength\tabcolsep{5.0pt}
	\begin{tabular}{lcc}
		\specialrule{0.4mm}{.0em}{.3em}
		Method & Backbone & mAP (\%)  \\
		\midrule
		3D CNN & I3D & 72.43 \\ 
		3D CNN + Timeception~\cite{hussein2018timeception} & I3D &  74.79 \\
		3D CNN + \textbf{PIC}  & I3D & \textbf{78.31} \\
		\specialrule{0.4mm}{.0em}{.0em}
	\end{tabular}
	\caption{PIC improves over related works in recognizing the multi-labeled, long-range videos of MultiThumos.}
	\label{tbl:4-7}
	\vspace*{-15pt}
\end{table}

\section{Conclusion}\label{sec:conclusions}
This paper introduces PIC, Permutation Invariant Convolution, a neural block dedicated to the temporal modeling of long-range activities in videos.
It has three properties.
First, being invariant to temporal permutations enable it to handle the chaotic temporal orders of long-range activities.
Second, it respects temporal locality, so it can learn deep temporal abstractions using a cascade of layers.
Third, it uses shared weights, namely key-value pairs, to learn the most representative visual signals in long and noisy videos.
We demonstrate the effectiveness of PIC layers, along with its three properties.
Most notably, we show how PIC enables existing CNNs to model long-range activities and improve the performance.
We benchmark on three datasets of long-range activities, where we improves on the previous methods.

{\small
\bibliographystyle{unsrt}
\bibliography{main}}

\begin{thebibliography}{10}

\bibitem{kuehne2014language}
Hilde Kuehne, Ali Arslan, and Thomas Serre.
\newblock The language of actions: Recovering the syntax and semantics of
  goal-directed human activities.
\newblock In {\em CVPR}, 2014.

\bibitem{hussein2018timeception}
Noureldien Hussein, Efstratios Gavves, and Arnold~WM Smeulders.
\newblock Timeception for complex action recognition.
\newblock In {\em CVPR}, 2019.

\bibitem{hussein2019videograph}
Noureldien Hussein, Efstratios Gavves, and Arnold~WM Smeulders.
\newblock Videograph: Recognizing minutes-long human activities in videos.
\newblock {\em ICCV Workhop}, 2019.

\bibitem{dushnik1941partially}
Ben Dushnik and Edwin~W Miller.
\newblock Partially ordered sets.
\newblock In {\em AJM}, 1941.

\bibitem{carreira2017quo}
Joao Carreira and Andrew Zisserman.
\newblock Quo vadis, action recognition? a new model and the kinetics dataset.
\newblock In {\em CVPR}, 2017.

\bibitem{wang2017non}
Xiaolong Wang, Ross Girshick, Abhinav Gupta, and Kaiming He.
\newblock Non-local neural networks.
\newblock In {\em CVPR}, 2018.

\bibitem{wu2019long}
Chao-Yuan Wu, Christoph Feichtenhofer, Haoqi Fan, Kaiming He, Philipp
  Krahenbuhl, and Ross Girshick.
\newblock Long-term feature banks for detailed video understanding.
\newblock In {\em CVPR}, 2019.

\bibitem{duta2017spatio}
Ionut~C Duta, Bogdan Ionescu, Kiyoharu Aizawa, and Nicu Sebe.
\newblock Spatio-temporal vlad encoding for human action recognition in videos.
\newblock In {\em ICMM}, 2017.

\bibitem{girdhar2017actionvlad}
Rohit Girdhar, Deva Ramanan, Abhinav Gupta, Josef Sivic, and Bryan Russell.
\newblock Actionvlad: Learning spatio-temporal aggregation for action
  classification.
\newblock In {\em CVPR}, 2017.

\bibitem{lea2017temporal}
Colin Lea, Michael~D Flynn, Rene Vidal, Austin Reiter, and Gregory~D Hager.
\newblock Temporal convolutional networks for action segmentation and
  detection.
\newblock In {\em CVPR}, 2017.

\bibitem{vaswani2017attention}
Ashish Vaswani, Noam Shazeer, Niki Parmar, Jakob Uszkoreit, Llion Jones,
  Aidan~N Gomez, {\L}ukasz Kaiser, and Illia Polosukhin.
\newblock Attention is all you need.
\newblock In {\em NIPS}, 2017.

\bibitem{sigurdsson2016hollywood}
Gunnar~A Sigurdsson, G{\'u}l Varol, Xiaolong Wang, Ali Farhadi, Ivan Laptev,
  and Abhinav Gupta.
\newblock Hollywood in homes: Crowdsourcing data collection for activity
  understanding.
\newblock In {\em ECCV}, 2016.

\bibitem{yeung2015every}
Serena Yeung, Olga Russakovsky, Ning Jin, Mykhaylo Andriluka, Greg Mori, and
  Li~Fei-Fei.
\newblock Every moment counts: Dense detailed labeling of actions in complex
  videos.
\newblock {\em International Journal of Computer Vision}, 2017.

\bibitem{soomro2012ucf101}
Khurram Soomro, Amir~Roshan Zamir, and Mubarak Shah.
\newblock Ucf101: A dataset of 101 human actions classes from videos in the
  wild.
\newblock In {\em CRCV-TR}, 2012.

\bibitem{karpathy2014large}
Andrej Karpathy, George Toderici, Sanketh Shetty, Thomas Leung, Rahul
  Sukthankar, and Li~Fei-Fei.
\newblock Large-scale video classification with convolutional neural networks.
\newblock In {\em CVPR}, 2014.

\bibitem{kay2017kinetics}
Will Kay, Joao Carreira, Karen Simonyan, Brian Zhang, Chloe Hillier, Sudheendra
  Vijayanarasimhan, Fabio Viola, Tim Green, Trevor Back, Paul Natsev, et~al.
\newblock The kinetics human action video dataset.
\newblock In {\em arXiv}, 2017.

\bibitem{ghodrati2018video}
Amir Ghodrati, Efstratios Gavves, and Cees~GM Snoek.
\newblock Video time: Properties, encoders and evaluation.
\newblock In {\em BMVC}, 2018.

\bibitem{wang2011action}
Heng Wang, Alexander Kl{\"a}ser, Cordelia Schmid, and Liu Cheng-Lin.
\newblock Action recognition by dense trajectories.
\newblock In {\em CVPR}, 2011.

\bibitem{jain2013better}
Mihir Jain, Herve Jegou, and Patrick Bouthemy.
\newblock Better exploiting motion for better action recognition.
\newblock In {\em CVPR}, 2013.

\bibitem{hussein2017unified}
Noureldien Hussein, Efstratios Gavves, and Arnold~WM Smeulders.
\newblock Unified embedding and metric learning for zero-exemplar event
  detection.
\newblock In {\em CVPR}, 2017.

\bibitem{damen2018scaling}
Dima Damen, Hazel Doughty, Giovanni Maria~Farinella, Sanja Fidler, Antonino
  Furnari, Evangelos Kazakos, Davide Moltisanti, Jonathan Munro, Toby Perrett,
  Will Price, et~al.
\newblock Scaling egocentric vision: The epic-kitchens dataset.
\newblock In {\em ECCV}, 2018.

\bibitem{idrees2017thumos}
Haroon Idrees, Amir~R Zamir, Yu-Gang Jiang, Alex Gorban, Ivan Laptev, Rahul
  Sukthankar, and Mubarak Shah.
\newblock The thumos challenge on action recognition for videos “in the
  wild”.
\newblock In {\em CVIU}, 2017.

\bibitem{Zhou2017YouCookIID}
Luowei Zhou and Jason~J. Corso.
\newblock Youcookii dataset.
\newblock In {\em arXiv}, 2017.

\bibitem{sener2019zero}
Fadime Sener and Angela Yao.
\newblock Zero-shot anticipation for instructional activities.
\newblock In {\em ICCV}, 2019.

\bibitem{krizhevsky2014imagenet}
Alex Krizhevsky, I~Sutskever, and G~Hinton.
\newblock Imagenet classification with deep convolutional neural.
\newblock In {\em NeurIPS}, 2014.

\bibitem{simonyan2014very}
Karen Simonyan and Andrew Zisserman.
\newblock Very deep convolutional networks for large-scale image recognition.
\newblock In {\em ICLR}, 2015.

\bibitem{he2016deep}
Kaiming He, Xiangyu Zhang, Shaoqing Ren, and Jian Sun.
\newblock Deep residual learning for image recognition.
\newblock In {\em CVPR}, 2016.

\bibitem{ji20123d}
Shuiwang Ji, Wei Xu, Ming Yang, and Kai Yu.
\newblock 3d convolutional neural networks for human action recognition.
\newblock {\em TPAMI}, 2012.

\bibitem{simonyan2014two}
Karen Simonyan and Andrew Zisserman.
\newblock Two-stream convolutional networks for action recognition in videos.
\newblock In {\em NIPS}, 2014.

\bibitem{gehring2017convolutional}
Jonas Gehring, Michael Auli, David Grangier, Denis Yarats, and Yann~N Dauphin.
\newblock Convolutional sequence to sequence learning.
\newblock In {\em ICML}, 2017.

\bibitem{szegedy2017inception}
Christian Szegedy, Sergey Ioffe, Vincent Vanhoucke, and Alexander~A Alemi.
\newblock Inception-v4, inception-resnet and the impact of residual connections
  on learning.
\newblock In {\em AAAI}, 2017.

\bibitem{xu2015show}
Kelvin Xu, Jimmy Ba, Ryan Kiros, Kyunghyun Cho, Aaron Courville, Ruslan
  Salakhudinov, Rich Zemel, and Yoshua Bengio.
\newblock Show, attend and tell: Neural image caption generation with visual
  attention.
\newblock In {\em ICML}, pages 2048--2057, 2015.

\bibitem{sharma2015action}
Shikhar Sharma, Ryan Kiros, and Ruslan Salakhutdinov.
\newblock Action recognition using visual attention.
\newblock {\em arXiv}, 2015.

\bibitem{du2018interaction}
Yang Du, Chunfeng Yuan, Bing Li, Lili Zhao, Yangxi Li, and Weiming Hu.
\newblock Interaction-aware spatio-temporal pyramid attention networks for
  action classification.
\newblock In {\em ECCV}, 2018.

\bibitem{li2018videolstm}
Zhenyang Li, Kirill Gavrilyuk, Efstratios Gavves, Mihir Jain, and Cees~GM
  Snoek.
\newblock Videolstm convolves, attends and flows for action recognition.
\newblock In {\em CVIU}, 2018.

\bibitem{velivckovic2017graph}
Petar Veli{\v{c}}kovi{\'c}, Guillem Cucurull, Arantxa Casanova, Adriana Romero,
  Pietro Lio, and Yoshua Bengio.
\newblock Graph attention networks.
\newblock In {\em ICLR}, 2018.

\bibitem{girdhar2019video}
Rohit Girdhar, Joao Carreira, Carl Doersch, and Andrew Zisserman.
\newblock Video action transformer network.
\newblock In {\em CVPR}, pages 244--253, 2019.

\bibitem{hussein2020timegate}
Noureldien Hussein, Mihir Jain, and Babak~Ehteshami Bejnordi.
\newblock Timegate: Conditional gating of segments in long-range activities.
\newblock In {\em arXiv}, 2020.

\bibitem{parmar2019stand}
Niki Parmar, Prajit Ramachandran, Ashish Vaswani, Irwan Bello, Anselm Levskaya,
  and Jonathon Shlens.
\newblock Stand-alone self-attention in vision models.
\newblock In {\em NeurIPS}, 2019.

\bibitem{lample2019large}
Guillaume Lample, Alexandre Sablayrolles, Marc'Aurelio Ranzato, Ludovic
  Denoyer, and Herv{\'e} J{\'e}gou.
\newblock Large memory layers with product keys.
\newblock In {\em arXiv}, 2019.

\bibitem{girdhar2017attentional}
Rohit Girdhar and Deva Ramanan.
\newblock Attentional pooling for action recognition.
\newblock In {\em NIPS}, 2017.

\bibitem{miech2017learnable}
Antoine Miech, Ivan Laptev, and Josef Sivic.
\newblock Learnable pooling with context gating for video classification.
\newblock In {\em arXiv}, 2017.

\bibitem{oneata2013action}
Dan Oneata, Jakob Verbeek, and Cordelia Schmid.
\newblock Action and event recognition with fisher vectors on a compact feature
  set.
\newblock In {\em ICCV}, 2013.

\bibitem{xie2017aggregated}
Saining Xie, Ross Girshick, Piotr Doll{\'a}r, Zhuowen Tu, and Kaiming He.
\newblock Aggregated residual transformations for deep neural networks.
\newblock In {\em CVPR}, 2017.

\bibitem{tensorflow2015-whitepaper}
Mart\'{\i}n Abadi et~al.
\newblock Tensorflow.
\newblock \url{tensorflow.org}, 2015.

\bibitem{chollet2015keras}
Fran\c{c}ois Chollet et~al.
\newblock Keras.
\newblock \url{keras.io}, 2015.

\bibitem{arandjelovic2013all}
Relja Arandjelovic and Andrew Zisserman.
\newblock All about vlad.
\newblock In {\em CVPR}, 2013.

\bibitem{feichtenhofer2019slowfast}
Christoph Feichtenhofer, Haoqi Fan, Jitendra Malik, and Kaiming He.
\newblock Slowfast networks for video recognition.
\newblock In {\em ICCV}, 2019.

\end{thebibliography}

\end{document}